\documentclass[conference]{IEEEtran}

\usepackage[utf8]{inputenc} 
\usepackage[T1]{fontenc}
\usepackage{url}
\usepackage{ifthen}
\usepackage{cite}
\usepackage[cmex10]{amsmath} 
\usepackage{amssymb}
\usepackage{mathtools}
\usepackage{bm}
\usepackage{color}
\usepackage{algorithm}
\usepackage{algpseudocode}

\newtheorem{exam}{Example}
\newtheorem{defi}{Definition}
\newtheorem{prop}{Proposition}
\newtheorem{theo}{Theorem}
\newtheorem{lemm}{Lemma}
\newtheorem{condi}{Condition}
\newtheorem{rema}{Remark}
\newtheorem{coro}{Corollary}

\newcommand{\argmax}{\mathop{\rm arg~max}\limits}

\begin{document}
\title{Probability Distribution on Rooted Trees} 



 \author{%
   \IEEEauthorblockN{Yuta Nakahara\IEEEauthorrefmark{1},
                     Shota Saito\IEEEauthorrefmark{2},
                     Akira Kamatsuka\IEEEauthorrefmark{3},
                     and Toshiyasu Matsushima\IEEEauthorrefmark{4}}
   \IEEEauthorblockA{\IEEEauthorrefmark{1}%
                     Center for Data Science, Waseda University, 
                     Tokyo, 169-8050, Japan, 
                     yuta.nakahara@aoni.waseda.jp}
   \IEEEauthorblockA{\IEEEauthorrefmark{2}%
                     Faculty of Informatics, Gunma University,
                     Gunma 371-8510, Japan
                     shota.s@gunma-u.ac.jp}
   \IEEEauthorblockA{\IEEEauthorrefmark{3}%
                     Dept. of Information Science, Shonan Institute of Technology,
                     Kanagawa, 251-8511, Japan
                     kamatsuka@info.shonan-it.ac.jp}
   \IEEEauthorblockA{\IEEEauthorrefmark{4}%
                     Dept. of Applied Math., Waseda University,
                     Tokyo, 169-8555, Japan,
                     toshimat@waseda.jp}
 }

\maketitle

\begin{abstract}
The hierarchical and recursive expressive capability of rooted trees is applicable to represent statistical models in various areas, such as data compression, image processing, and machine learning. On the other hand, such hierarchical expressive capability causes a problem in tree selection to avoid overfitting. One unified approach to solve this is a Bayesian approach, on which the rooted tree is regarded as a random variable and a direct loss function can be assumed on the selected model or the predicted value for a new data point. However, all the previous studies on this approach are based on the probability distribution on full trees, to the best of our knowledge. In this paper, we propose a generalized probability distribution for any rooted trees in which only the maximum number of child nodes and the maximum depth are fixed. Furthermore, we derive recursive methods to evaluate the characteristics of the probability distribution without any approximations.
\end{abstract}

\section{Introduction}\label{introduction}

The hierarchical and recursive expressive capability of rooted trees is utilized in various fields of study. They serve as an index of a statistical model or function, i.e., one rooted tree $\tau$ corresponds to one statistical model $p(x; \tau)$ or one function $f_\tau (x)$. For example, for text compression in information theory, a rooted tree represents a set of contexts, which are strings of the most recent symbols and govern the probabilistic generation of the next symbol at each time point. This tree is known as a context tree\cite{CTW, CT_th, CT_alg, Papageorgiou, kontoyiannis}. In image processing, a rooted tree represents a procedure to capture non-stationarity among variable size block regions, and it is known as quadtree block partitioning\cite{H265, nakahara_entropy}. In machine learning, a rooted tree represents a nonlinear function that comprises many conditional branches and is known as a decision tree\cite{CART, RF, XGBoost, meta-tree}.

However, such hierarchical expressive capability causes difficulty in tree selection, i.e., the selection of one statistical model or function. Since the deeper tree hierarchically contains the shallow one, the most likely tree for given data is inevitably the deepest one. This results in losing the consistency of the estimated model or deteriorating the prediction accuracy for a new data point.\footnote{Such a phenomenon is called ``overfitting'' in the field of machine learning at times.}

Approaches to this difficulty are divided into two types. The first one is a non-Bayesian approach. On this approach, previous studies regarded the rooted trees as unknown constants. They have provided algorithmic modifications of the tree selection, e.g., applying a stopping rule for node expansion\cite{H265, CART}, introducing a normalization term into the objective function\cite{XGBoost}, or averaging the statistical models or the functions with some weights\cite{CTW, RF, XGBoost}. However, these algorithmic modifications are heuristic at times. Model selection criteria such as Akaike's information criterion (AIC) \cite{AIC} can also be applied if the tree corresponds to a statistical model.

The second one is a Bayesian approach. On this approach, previous studies regarded the rooted trees as a random variable and assumed a prior distribution, which provided a unified solution to the difficulty in the tree selection. They could directly assume a loss function for the estimated tree or the predicted value for a new data point based on the Bayes decision theory (see, e.g., \cite{Berger}).\footnote{Although the Bayes decision theory is typically applied to statistical models with unknown continuous parameters, it is also applicable to statistical models with unknown discrete random variables such as rooted trees (see, e.g., \cite{Bayes_code}).} This enables selecting one rooted tree or combining them according to the posterior distribution. In particular, deeper trees will be avoided by assigning a high prior probability to a shallow tree and a low prior probability to a deep tree. In terms of text compression, the complete Bayesian interpretation of the context tree weighting method was investigated by the authors of \cite{CT_th, CT_alg}. Moreover, similar results obtained from rich real data analysis have been reported recently \cite{Papageorgiou, kontoyiannis}.\footnote{Note that the prior form reported in \cite{Papageorgiou, kontoyiannis} is restricted and cannot be updated as a posterior, in contrast to that reported in \cite{CT_th, CT_alg}.} In image processing, the authors of \cite{nakahara_entropy} regarded the quadtree as a stochastic model and optimally estimated it under the Bayes criterion. In machine learning, the authors of \cite{meta-tree} redefined the decision tree as a stochastic generative model to improve various tree weighting methods (e.g., \cite{RF}). The mathematically essential part of these studies was summarized in \cite{full_rooted_tree_arxiv}. However, all these studies are based on a probability distribution on full trees, i.e., the rooted trees whose inner nodes have the same number of children.

In this paper, we adopt the second approach and propose a generalized probability distribution on any rooted trees in which only the maximum number of child nodes and the maximum depth are fixed. Consequently, we derive recursive methods to evaluate the characteristics of the probability distribution on rooted trees. They enable us to calculate marginal distributions for each node, the mode of the tree distribution, expectations of some class of functions, and the posterior distribution for a class of likelihoods, without any approximations. Although the computational complexity of our methods exponentially increases with respect to the maximum number of child nodes, this is not so problematic in some practical situations. An example of applications will be described in Section \ref{application}.

The remainder of this paper is organized as follows: In Section \ref{sec-notations}, we present the notations used herein. In Section \ref{sec-definition}, we define the prior on rooted trees. In Section \ref{sec-properties}, we describe the algorithms for calculating the properties of the proposed distribution, e.g., a marginal distribution for each node, an efficient calculation of the expectation, mode, and the posterior distribution. In Section \ref{discussion}, we discuss the usefulness of our distribution in statistical decision theory and hierarchical Bayesian modeling. In Section \ref{application}, we describe an example of applications of our probability distribution. In Section \ref{sec-future}, we propose some future work. In Section \ref{sec-conclusion}, we conclude the paper.

\section{Notations used for rooted trees}\label{sec-notations}

\begin{figure}[tbp]
  \centering
  \includegraphics[width=0.45\textwidth]{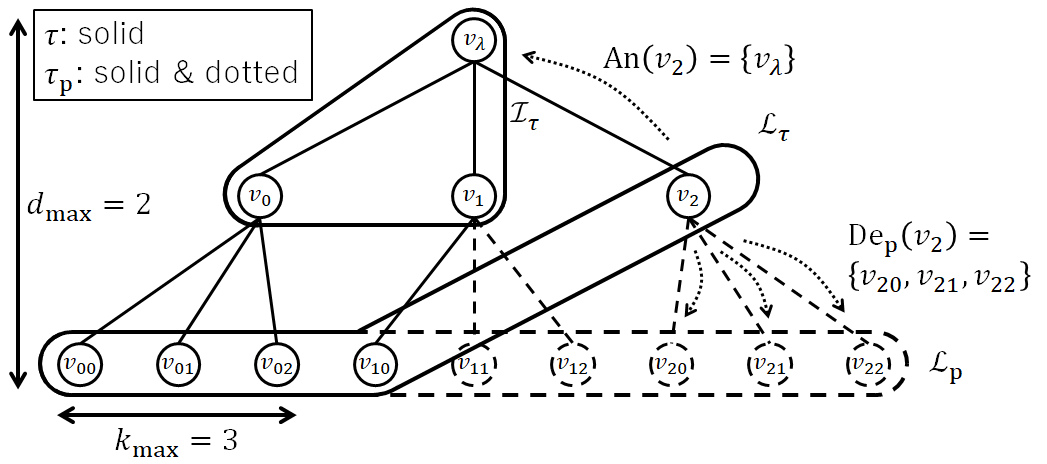}
  \caption{The notations for the rooted trees.}
  \label{notations}
\end{figure}

In this section, we define notations for the rooted trees. They are shown in Fig. \ref{notations}. Let $k_\mathrm{max} \in \mathbb{N}$ denote the maximum number of child nodes and $d_\mathrm{max} \in \mathbb{N}$ denote the maximum depth. Let $\tau_\mathrm{p} = (\mathcal{V}_\mathrm{p}, \mathcal{E}_\mathrm{p})$ denote the perfect\footnote{``Perfect'' means that all inner nodes have exactly $k_\mathrm{max}$ children and all leaf nodes have the same depth.} $k_\mathrm{max}$-ary rooted tree whose depth is $d_\mathrm{max}$ and root node is $v_\lambda$. $\mathcal{V}_\mathrm{p}$ and $\mathcal{E}_\mathrm{p}$ denote the set of the nodes and edges of it, respectively. Then, let $\mathcal{I}_\mathrm{p} \subset \mathcal{V}_\mathrm{p}$ and $\mathcal{L}_\mathrm{p} \subset \mathcal{V}_\mathrm{p}$ denote the set of the inner nodes and the leaf nodes of $\tau_\mathrm{p}$, respectively. For each node $v \in \mathcal{V}_\mathrm{p}$, $\mathrm{Ch}_\mathrm{p}(v) \subset \mathcal{V}_\mathrm{p}$ denote the set of child nodes of $v$ on $\tau_\mathrm{p}$, and $v_\mathrm{pa}$ denote the parents node of $v$. Notations used for the relation between two nodes $v, v' \in \mathcal{V}_\mathrm{p}$ are as follows. Let $v \succ v'$ denote that $v$ is an ancestor node of $v'$, ($v'$ is a descendant node of $v$), $v \succeq v'$ denote that $v$ is an ancestor node of $v'$ or $v'$ itself, ($v'$ is a descendant node of $v$ or $v$ itself), $\mathrm{An}(v) \coloneqq \{ v' \in \mathcal{V}_\mathrm{p} \mid v' \succ v \}$, and $\mathrm{De}_\mathrm{p} (v) \coloneqq \{ v' \in \mathcal{V}_\mathrm{p} \mid v' \prec v \}$.

Subsequently, we consider rooted subtrees of $\tau_\mathrm{p}$ in which their root nodes are the same as $v_\lambda$. Let $\mathcal{T}$ denote the set of all rooted subtrees of $\tau_\mathrm{p}$. They are called rooted subtrees and $\tau_\mathrm{p}$ is called the base tree. Let $\mathcal{V}_\tau$ and $\mathcal{E}_\tau$ denote the set of the nodes and the edges of $\tau \in \mathcal{T}$, respectively. Let $\mathcal{I}_\tau \subset \mathcal{V}_\tau$ and $\mathcal{L}_\tau \subset \mathcal{V}_\tau$ denote the set of the inner nodes and the leaf nodes of $\tau \in \mathcal{T}$, respectively. Lastly, let $\mathcal{E}_{v \to v'}$ denote edges on the path from $v$ to $v'$.

\section{Definition of probability distribution on rooted subtrees}\label{sec-definition}

In this section, we define a probability distribution on rooted subtrees $\mathcal{T}$. Let $T$ denote the random variable on $\mathcal{T}$, and $\tau$ denote its realization.

\begin{defi}\label{edge_spreading_pattern}
For $\tau \in \mathcal{T}$, we define a vector $\bm z_v^\tau \in \{ 0, 1\}^{k_\mathrm{max}}$ representing an edge spreading pattern of $v$ in $\mathcal{V}_\tau$ as
\begin{align}
\bm z_v^\tau \coloneqq
\begin{cases}
(z_{vv'}^\tau)_{v' \in \mathrm{Ch}_\mathrm{p}(v)} \coloneqq (I\{ v' \in \mathcal{V}_\tau \})_{v' \in \mathrm{Ch}_\mathrm{p} (v)}, & v \in \mathcal{I}_\mathrm{p}, \\
\bm 0, & v \in \mathcal{L}_\mathrm{p},
\end{cases}
\end{align}
where $I \{ \cdot \}$ denotes the indicator function.
\end{defi}

\begin{figure}[tbp]
  \centering
  \includegraphics[width=0.4\textwidth]{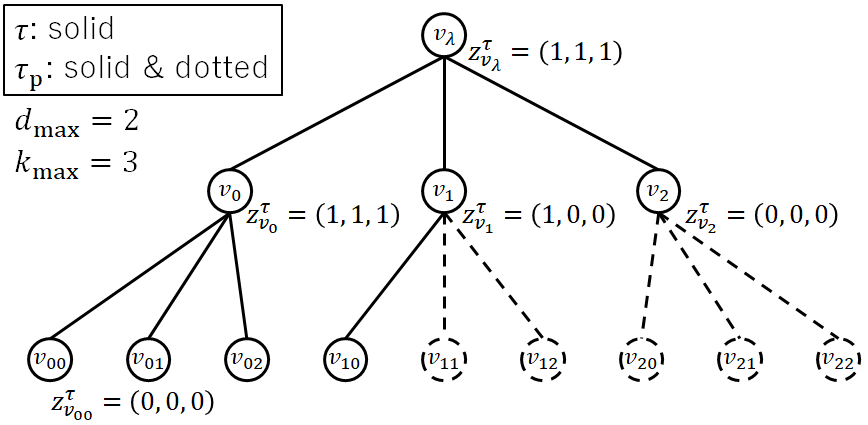}
  \caption{
  Examples of $\bm z_v^\tau$ for some nodes $v$ in $\tau$, which is shown with the solid lines. Here, $k_\mathrm{max} = 3$ and $d_\mathrm{max} = 2$. 
  }
  \label{example_of_z}
\end{figure}

\begin{exam}
Figure \ref{example_of_z} shows examples of $\bm z_v^\tau$.
\end{exam}

\begin{defi}\label{general_def}
Let $\theta_v (\bm z) \in [0, 1]$ be a given hyperparameter (a mapping from $\{ 0, 1\}^{k_\mathrm{max}}$ to $[0, 1]$) of a node $v \in \mathcal{V}_\mathrm{p}$, which satisfies $\sum_{\bm z \in \{ 0, 1\}^{k_\mathrm{max}}} \theta_v (\bm z) = 1$. Then, we define a probability distribution on $\mathcal{T}$ as follows.
\begin{align}
p(\tau) \coloneqq \prod_{v \in \mathcal{V}_\tau} \theta_v (\bm z_v^\tau) \text{ for all } \tau \in \mathcal{T}, \label{definition_of_distribution}
\end{align}
where $\theta_v (\bm 0) = 1$ for $v \in \mathcal{L}_\mathrm{p}$.
\end{defi}

Intuitively, $\theta_v (\bm z_v^\tau)$ represents the probability that $v$ has the edge spreading pattern $\bm z_v^\tau$ under the condition that $v$ is contained in the tree $\tau$.\footnote{It will be proved as a theoretical fact in Remark \ref{conditional}.} Therefore, the occurrence probability of a rooted subtree exponentially decays as its depth increases.

\begin{figure*}[tbp]
  \centering
  \includegraphics[width=0.8\textwidth]{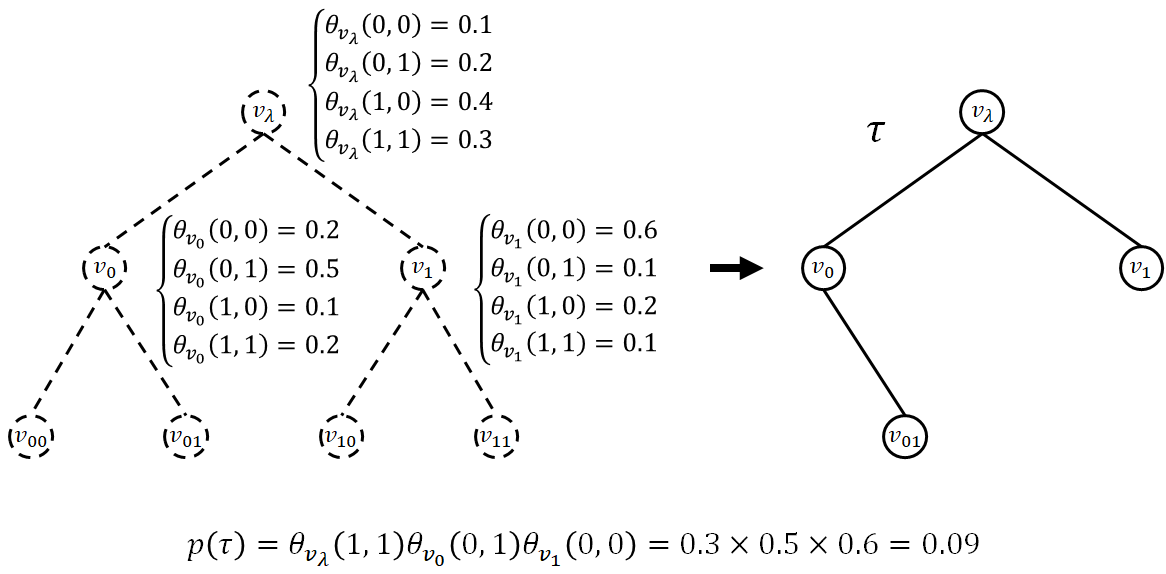}
  \caption{
  An example of the probability that the subtree $\tau$ shown in right-hand side is generated from the base tree with the parameters shown in the left. Here, $k_\mathrm{max} = 2$ and $d_\mathrm{max} = 2$.
  }
  \label{example}
\end{figure*}

\begin{exam}
The right-hand side of Fig.\ \ref{example} shows an example of a probability that a tree occurs under given hyperparameters in the left side of Fig.\ \ref{example}.
\end{exam}

\begin{rema}
Definition \ref{general_def} contains the definition of the probability distribution for the full trees in \cite{full_rooted_tree_arxiv} as a specific case when there exists $\alpha_v \in [0, 1]$ for each $v \in \mathcal{I}_\mathrm{p}$ such that
\begin{align}
\theta_v (\bm z) =
\begin{cases}
\alpha_v, & \bm z = (1, 1, \dots , 1), \\
1-\alpha_v, & \bm z = \bm 0, \\
0, & \mathrm{otherwise}.
\end{cases}
\end{align}
\end{rema}

\begin{theo}\label{sum1}
The quantity $p(\tau)$ defined as in (\ref{definition_of_distribution}) fulfills the condition of the probability distribution, that is, $\sum_{\tau \in \mathcal{T}} p(\tau) = 1$.
\end{theo}

We will first prove Lemma \ref{func_sum}, which is the essential lemma since it is not used only in the proof of Theorem \ref{sum1} but also in the proof of other theorems later.

\begin{lemm}\label{func_sum}
Let $F: \mathcal{T} \to \mathbb{R}$ be a real-valued function on the set $\mathcal{T}$ of the rooted subtrees of the base tree $\tau_\mathrm{p}$. If $F$ has the form
\begin{align}
F(\tau) = \prod_{v \in \mathcal{V}_\tau} G_v(\bm z_v^\tau) \text{ for all } \tau \in \mathcal{T},
\end{align}
where $G_v: \{ 0, 1\}^{k_\mathrm{max}} \to \mathbb{R}$ is real-valued functions on $\{ 0, 1\}^{k_\mathrm{max}}$ for each $v \in \mathcal{V}_\mathrm{p}$, then the summation $\sum_{\tau \in \mathcal{T}} F(\tau)$ can be recursively decomposed as follows.
\begin{align}
\sum_{\tau \in \mathcal{T}} F(\tau) = \phi (v_\lambda),
\end{align}
where $\phi: \mathcal{V}_\mathrm{p} \to \mathbb{R}$ is defined as below.
\begin{align}
\phi (v) \coloneqq
\begin{cases}
G_v(\bm 0), & v \in \mathcal{L}_\mathrm{p},\\
\sum_{\bm z_v \in \{ 0, 1\}^{k_\mathrm{max}}} \Bigl\{ G_v(\bm z_v) \\
\qquad \times \prod_{v' \in \mathrm{Ch}_\mathrm{p} (v)} \left( \phi (v') \right)^{z_{vv'}} \Bigr\} , & v \in \mathcal{I}_\mathrm{p},
\end{cases}\label{phi_v}
\end{align}
where $z_{vv'}$ denotes an element of $\bm z_v \in \{ 0, 1\}^{k_\mathrm{max}}$, which corresponds to $v' \in \mathrm{Ch}_\mathrm{p} (v)$. Note that $\bm z_v$ is the local variable defined in the summation and independent of $\tau$. Therefore, we denote it without $\tau$. Similar notations will be used throughout this paper.
\end{lemm}

\textit{Proof:} The cases of the sum is divided as follows.
\begin{align}
&\sum_{\tau \in \mathcal{T}} F(\tau) \nonumber \\
&= \sum_{\tau \in \mathcal{T}} \prod_{v \in \mathcal{V}_\tau} G_v(\bm z_v^\tau) \label{root_sum} \\
&= \sum_{\bm z_{v_\lambda} \in \{ 0, 1\}^{k_\mathrm{max}}} \sum_{\tau \in \mathcal{T} : \bm z_{v_\lambda}^\tau = \bm z_{v_\lambda}} \prod_{v \in \mathcal{V}_\mathrm{p}} G_v (\bm z_v^\tau) \\
&= \sum_{\bm z_{v_\lambda} \in \{ 0, 1\}^{k_\mathrm{max}}} G_{v_\lambda} (\bm z_{v_\lambda}) \sum_{\tau \in \mathcal{T} : \bm z_{v_\lambda}^\tau = \bm z_{v_\lambda}} \prod_{v \in \mathcal{V}_\mathrm{p} \setminus \{ v_\lambda \}} G_v (\bm z_v^\tau). \label{factorization}
\end{align}

\begin{figure}[tbp]
  \centering
  \includegraphics[width=0.45\textwidth]{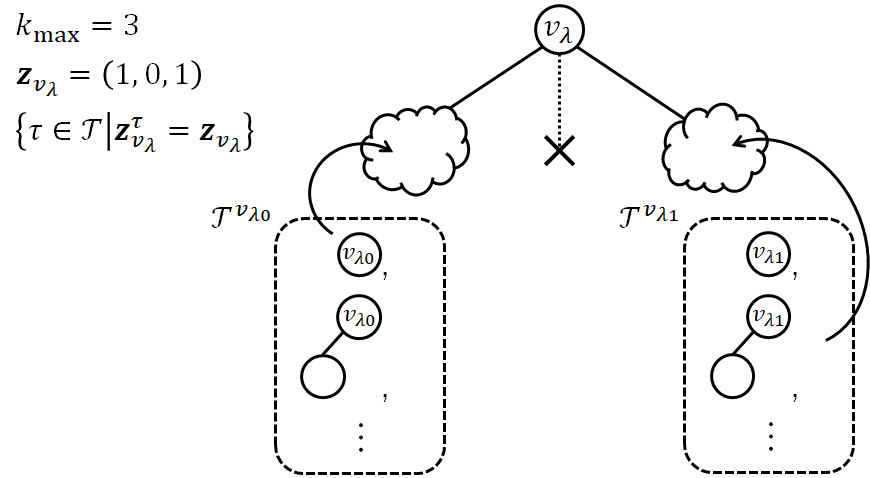}
  \caption{
  Structure of the trees in $\{ \tau \in \mathcal{T} | \bm z_{v_\lambda}^\tau = \bm z_{v_\lambda} \}$, where $k_\mathrm{max} = 3$ and $\bm z_{v_\lambda} = (1,0,1)$. All of them have the root node $v_\lambda$ whose edge spreading pattern is $\bm z_{v_\lambda} = (1, 0, 1)$. The other structure is determined by choosing subtrees from $\mathcal{T}^{v_{\lambda 0}}$ and $\mathcal{T}^{v_{\lambda 1}}$.
  }
  \label{factorization_img}
\end{figure}

We focus on the tree $\tau$ whose root node $v_\lambda$ has the edge spreading pattern $\bm z_{v_\lambda}^\tau = \bm z_{v_\lambda}$. Let $k$ denote the number of edges emitted from $v_\lambda$, that is, $k \coloneqq \sum_{v \in \mathrm{Ch}_\mathrm{p}(v_\lambda)} z_{v_\lambda v}$. We index them in an appropriate order, and let $v_{\lambda i}$ denote the $i$-th child node of $v_\lambda$ for $i \in \{ 0, 1, \dots , k-1\}$, which satisfies $z_{v_\lambda v_{\lambda i}} = 1$. The other structure of $\tau$ is determined by the shape of $k$ subtrees whose root nodes are $(v_{\lambda i})_{i \in \{ 0, \dots , k-1\}}$ (see Fig. \ref{factorization_img}). Let $\mathcal{T}^{v_{\lambda i}}$ denote the set of subtrees whose root node is $v_{\lambda i}$. Then, there is a natural bijection from $\{ \tau \in \mathcal{T} | \bm z_{v_\lambda}^\tau = \bm z_{v_\lambda} \}$ to the product set $\mathcal{T}^{v_{\lambda 0}} \times \cdots \times \mathcal{T}^{v_{\lambda \, k-1}}$. Therefore, the summation of (\ref{factorization}) is further factorized. Consequently, we have

\begin{align}
&(\ref{factorization}) = \sum_{\bm z_{v_\lambda} \in \{ 0, 1\}^{k_\mathrm{max}}} \Biggl\{ G_{v_\lambda} (\bm z_{v_\lambda}) \nonumber \\
&\times \sum_{(\tau_0, \dots , \tau_{k\!-\!1}) \in \mathcal{T}^{v_{\lambda 0}} \times \cdots \times \mathcal{T}^{v_{\lambda \, k\!-\!1}}} \Biggl[ \prod_{v \in \mathcal{V}_{\tau_0}} G_v(\bm z_{v}^{\tau_0}) \nonumber \\
&\qquad \qquad \qquad \qquad \qquad \qquad \cdots \prod_{v \in \mathcal{V}_{\tau_{k\!-\!1}}} G_v(\bm z_v^{\tau_{k\!-\!1}}) \Biggr] \Biggr\} \\
&= \sum_{\bm z_{v_\lambda} \in \{ 0, 1\}^{k_\mathrm{max}}} \Biggl\{ G_{v_\lambda} (\bm z_{v_\lambda}) \nonumber \\
&\times \sum_{\tau_0 \in \mathcal{T}^{v_{\lambda 0}}} \! \cdots \! \sum_{\tau_{k\!-\!1} \in \mathcal{T}^{v_{\lambda \, k\!-\!1}}} \Biggl[ \prod_{v \in \mathcal{V}_{\tau_0}} G_v(\bm z_{v}^{\tau_0}) \nonumber \\
&\qquad \qquad \qquad \qquad \qquad \qquad \cdots \prod_{v \in \mathcal{V}_{\tau_{k\!-\!1}}} G_v(\bm z_v^{\tau_{k\!-\!1}}) \Biggr] \Biggr\} \\
&= \sum_{\bm z_{v_\lambda} \in \{ 0, 1\}^{k_\mathrm{max}}} G_{v_\lambda} (\bm z_{v_\lambda}) \prod_{i=0}^{k-1} \sum_{\tau \in \mathcal{T}^{v_{\lambda i}}} \prod_{v \in \mathcal{V}_{\tau_i}} G_v(\bm z_{v}^{\tau_i}). \label{using_k}
\end{align}

To unify the notation, let $\mathcal{T}^{v}$ denote the set of subtrees whose root node is $v \in \mathcal{V}_\mathrm{p}$ in general, i.e., we define a notion similar to $\mathcal{T}^{v_{\lambda i}}$ for not only $v_{\lambda 0}, v_{\lambda 1}, \dots , v_{\lambda \, k-1}$ but also any other nodes $v \in \mathcal{V}_\mathrm{p}$. By using this notation, we have
\begin{align}
&(\ref{using_k}) = \nonumber \\
&\sum_{\bm z_{v_\lambda} \in \{ 0, 1\}^{k_\mathrm{max}}} \! \! G_{v_\lambda} (\bm z_{v_\lambda}) \! \! \prod_{v \in \mathrm{Ch}_\mathrm{p}(v_\lambda)} \! \left( \sum_{\tau \in \mathcal{T}^v} \prod_{v' \in \mathcal{V}_\tau} G_{v'} (\bm z_{v'}^\tau) \right)^{\! z_{v_\lambda v}}. \label{child_sum}
\end{align}

Then, from (\ref{root_sum}) and (\ref{child_sum}), we have
\begin{align}
&\underbrace{\sum_{\tau \in \mathcal{T}} \prod_{v \in \mathcal{V}_\tau} G_v(\bm z_v^\tau)}_{(a)} = \nonumber \\
& \sum_{\bm z_{v_\lambda} \in \{ 0, 1\}^{k_\mathrm{max}}} \! \! G_{v_\lambda} (\bm z_{v_\lambda}) \! \! \prod_{v \in \mathrm{Ch}_\mathrm{p}(v_\lambda)} \! \Biggl( \underbrace{\sum_{\tau \in \mathcal{T}^v} \prod_{v' \in \mathcal{V}_\tau} G_{v'} (\bm z_{v'}^\tau)}_{(b)} \Biggr)^{z_{v_\lambda v}}. \label{recursive_structure}
\end{align}
The underbraced parts $(a)$ and $(b)$ have the same structure except for the depth of the root node of the subtree. Therefore, $(b)$ can be decomposed in a similar manner from (\ref{root_sum}) to (\ref{child_sum}). We can continue this decomposition to the leaf nodes $v_\mathrm{leaf} \in \mathcal{L}_\mathrm{p}$, where $G_{v_{\mathrm{leaf}}} (\bm z_{v_\mathrm{leaf}}) = G_{v_\mathrm{leaf}}(\bm 0)$. Finally, we have an alternative definition of $\phi (v): \mathcal{V}_{\mathrm{p}} \to \mathbb{R}$, which is equivalent to (\ref{phi_v}).
\begin{align}
\phi (v) \coloneqq \sum_{\tau \in \mathcal{T}^v} \prod_{v \in \mathcal{V}_\tau} G_v (\bm z_v^\tau).
\end{align}
The equivalence is confirmed by substituting it into both sides of (\ref{recursive_structure}). Therefore, Lemma \ref{func_sum} is proved. \hfill $\blacksquare$

Then, the proof of Theorem \ref{sum1} is as follows.

\textit{Proof:} Using Lemma \ref{func_sum}, we can divide the cases of the sum and factorize the common terms of $\sum_{\tau \in \mathcal{T}} p(\tau)$ in the following recursive manner.
\begin{align}
\sum_{\tau \in \mathcal{T}} p(\tau) = \phi (v_\lambda),
\end{align}
where
\begin{align}
\phi (v) \coloneqq
\begin{cases}
\theta_v(\bm 0), & v \in \mathcal{L}_\mathrm{p},\\
\sum_{\bm z_v \in \{ 0, 1\}^{k_\mathrm{max}}} \Bigl\{ \theta_v(\bm z_v) \\
\qquad \times \prod_{v' \in \mathrm{Ch}_\mathrm{p} (v)} \left( \phi (v') \right)^{z_{vv'}} \Bigr\}, & v \in \mathcal{I}_\mathrm{p}.
\end{cases} \label{sum_recursive}
\end{align}

Then, we prove $\phi (v) = 1$ for any node $v \in \mathcal{V}_\mathrm{p}$ by structural induction. For any leaf node $v \in \mathcal{L}_\mathrm{p}$, $\theta_v (\bm 0) = 1$ from Definition \ref{definition_of_distribution}. Therefore,
\begin{align}
\phi (v) = \theta_v (\bm 0) = 1, \qquad v \in \mathcal{L}_\mathrm{p}.
\end{align}

For any inner node $v \in \mathcal{I}_\mathrm{p}$, assuming $\phi (v') = 1$ as the induction hypothesis for any descendant nodes $v' \in \mathrm{De}_\mathrm{p} (v)$,
\begin{align}
\phi (v) &= \sum_{\bm z_v \in \{ 0, 1\}^{k_\mathrm{max}}} \theta_v(\bm z_v) \prod_{v' \in \mathrm{Ch}_\mathrm{p} (v)} \left( \phi (v') \right)^{z_{vv'}} \\
&= \sum_{\bm z_v \in \{ 0, 1\}^{k_\mathrm{max}}} \theta_v(\bm z_v) \prod_{v' \in \mathrm{Ch}_\mathrm{p} (v)} 1^{z_{vv'}} \\
&= \sum_{\bm z_v \in \{ 0, 1\}^{k_\mathrm{max}}} \theta_v(\bm z_v) \\
&= 1.
\end{align}
The last equation is because the assumption of $\theta_v (\bm z)$ described in Definition \ref{definition_of_distribution}. Therefore, $\sum_{\tau \in \mathcal{T}} p(\tau) = \phi (v_\lambda) = 1$ since $v_\lambda$ is also in $\mathcal{V}_\mathrm{p}$. \hfill $\blacksquare$

\section{Properties of probability distribution on rooted subtrees}\label{sec-properties}

In this section, we describe properties of the probability distribution on rooted subtrees and methods to calculate them. All the proofs are in Appendix \ref{appendix}. Note that the motivation and usefulness of Conditions \ref{product}, \ref{sum}, \ref{generative-general}, and \ref{generative-path} in this section will be described in Sections \ref{discussion} and \ref{application}.

\subsection{Probability of events on nodes and edges}
At the beginning, we explain why $(\bm z_v^T = \bm z) \land (v \in \mathcal{V}_T)$ determines a probabilistic event. We consider any $\bm z \in \{ 0, 1\}^{k_\mathrm{max}}$ and $v \in \mathcal{V}_\mathrm{p}$ are given as non-stochastic constants and fixed. After that, a rooted subtree is randomly chosen according to the probability distribution proposed in Section \ref{sec-definition}. Then, the random variable $\bm z_v^T$ sometimes satisfies $\bm z_v^T = \bm z$ and sometimes not, depending on the realization $\tau$ of random variable $T$. Similarly, $\mathcal{V}_T$ sometimes contains $v$ and sometimes not, depending on its realization $\tau$ of random variable $T$. Therefore, $(\bm z_v^T = \bm z) \land (v \in \mathcal{V}_T)$ determines a probabilistic event on $p(\tau)$. Although the probability of this event is trivially represented as $\sum_{\tau \in \mathcal{T}} I \{ (\bm z_v^\tau = \bm z) \land (v \in \mathcal{V}_\tau) \} p(\tau)$, where $I \{ \cdot \}$ denotes the indicator function, we derive a computationally efficient form without the summation about $\tau$ in the following.

\begin{theo}\label{node_events}
For any $\bm z \in \{ 0, 1\}^{k_\mathrm{max}}$ and $v \in \mathcal{V}_\mathrm{p}$, we have the following:
\begin{align}
&\mathrm{Pr} \{ (\bm z_v^T = \bm z) \land (v \in \mathcal{V}_T) \} \nonumber \\
&= \theta_v(\bm z) \prod_{(v',v'') \in \mathcal{E}_{v_\lambda \to v}} \sum_{\bm z_{v'}: z_{v'v''} = 1} \theta_{v'} (\bm z_{v'}), \label{any_node_z_eq}
\end{align}
where $\mathcal{E}_{v_\lambda \to v}$ denotes edges on the path from $v_\lambda$ to $v$.
\end{theo}

\begin{rema}\label{conditional}
In the following, $v_\mathrm{pa}$ denotes the parents node of $v$. Probabilities of many other events on nodes and edges are derived from Theorem \ref{node_events}. For example,
\begin{align}
&\mathrm{Pr} \{ v \in \mathcal{V}_T \} \nonumber \\
&\quad = \sum_{\bm z \in \{ 0, 1\}^{k_\mathrm{max}}} \mathrm{Pr} \{ (\bm z_v^T = \bm z) \land (v \in \mathcal{V}_T) \} \\
&\quad = \sum_{\bm z \in \{ 0, 1\}^{d_\mathrm{max}}} \theta_v(\bm z) \prod_{(v',v'') \in \mathcal{E}_{v_\lambda \to v}} \sum_{\bm z_{v'}: z_{v'v''} = 1} \theta_{v'} (\bm z_{v'})\\
&\quad = \prod_{(v',v'') \in \mathcal{E}_{v_\lambda \to v}} \sum_{\bm z_{v'}: z_{v'v''} = 1} \theta_{v'} (\bm z_{v'}), \label{any_node_eq}\\
&\mathrm{Pr} \{ ( (v_\mathrm{pa}, v) \in \mathcal{E}_T) \land (v_\mathrm{pa} \in \mathcal{V}_T) \} \nonumber \\
&\quad = \sum_{\bm z_{v_\mathrm{pa}}: z_{v_\mathrm{pa} v}=1} \mathrm{Pr} \{ (\bm z_{v_\mathrm{pa}}^T = \bm z_{v_\mathrm{pa}}) \land (v_\mathrm{pa} \in \mathcal{V}_T) \} \\
&\quad = \sum_{\bm z_{v_\mathrm{pa}}: z_{v_\mathrm{pa} v}=1} \! \! \theta_{v_\mathrm{pa}}(\bm z_{v_\mathrm{pa}}) \! \! \prod_{(v',v'') \in \mathcal{E}_{v_\lambda \to v_\mathrm{pa}}} \sum_{\bm z_{v'}: z_{v'v''} = 1} \! \! \theta_{v'} (\bm z_{v'})\\
&\quad = \prod_{(v',v'') \in \mathcal{E}_{v_\lambda \to v_\mathrm{pa}}} \sum_{\bm z_{v'}: z_{v'v''} = 1} \theta_{v'} (\bm z_{v'}), \label{any_edge_eq}\\
&\mathrm{Pr} \{ v \in \mathcal{L}_T \} \nonumber \\
&\quad = \theta_v (\bm 0) \mathrm{Pr} \{ v \in \mathcal{V}_T \}, \label{leaf_node_eq}\\
&\mathrm{Pr} \{ v \in \mathcal{I}_T \} \nonumber \\
&\quad = \mathrm{Pr} \{ v \in \mathcal{V}_T \} - \mathrm{Pr} \{ v \in \mathcal{L}_T \} \\
&\quad = (1 - \theta_v (\bm 0)) \mathrm{Pr} \{ v \in \mathcal{V}_T \}, \label{inner_node_eq}\\
&\mathrm{Pr} \{ \bm z_v^T = \bm z \mid v \in \mathcal{V}_T \} \nonumber \\
&\quad = \frac{\mathrm{Pr} \{ (\bm z_v^T = \bm z) \land (v \in \mathcal{V}_T) \}}{\mathrm{Pr} \{ v \in \mathcal{V}_T \}} \\
&\quad = \theta_v (\bm z), \\
&\mathrm{Pr} \{ (v_\mathrm{pa}, v) \in \mathcal{E}_T \mid v_\mathrm{pa} \in \mathcal{V}_T \} \nonumber \\
&\quad = \frac{\mathrm{Pr} \{ ((v_\mathrm{pa}, v) \in \mathcal{E}_T) \land (v_\mathrm{pa} \in \mathcal{V}_T) \}}{\mathrm{Pr} \{ v_\mathrm{pa} \in \mathcal{V}_T \}} \\
&\quad = \sum_{\bm z_{v_\mathrm{pa}}: z_{v_\mathrm{pa} v}=1} \theta_{v_\mathrm{pa}}(\bm z_{v_\mathrm{pa}}).
\end{align}

\end{rema}

\subsection{Mode}
We describe an algorithm to find the mode of $p(\tau)$ with $O \left( 2^{k_\mathrm{max}} k_\mathrm{max}^{d_\mathrm{max}+1} \right)$ computational cost.\footnote{$O (\cdot)$ denotes the Big-O notation, i.e., $f(n) = O(g(n))$ means that $^\exists k > 0, ^\exists n_0 > 0, ^\forall n > n_0, |f(n)| \leq k \cdot g(n)$.} Note that, the size of search space $\mathcal{T}$ is of the order of $\Omega \left( 2^{k_\mathrm{max}^{d_\mathrm{max}-1}} \right)$ in general.\footnote{$\Omega ( \cdot )$ denotes the Big-Omega notation in complexity theory, i.e., $f(n) = \Omega(g(n))$ means that $^\exists k > 0, ^\exists n_0 > 0, ^\forall n > n_0, f(n) \geq k \cdot g(n)$. $|\mathcal{T}| = \Omega \left( 2^{k_\mathrm{max}^{d_\mathrm{max}-1}} \right)$ is proved by substituting $G_v(\bm z) \equiv 1$ in Lemma \ref{func_sum}.} First, replacing all the sum in the proof of Lemma \ref{func_sum} for the max, we can derive the following recursive expression of $\max_{\tau \in \mathcal{T}} p(\tau)$.

\begin{prop}
\begin{align}
\max_{\tau \in \mathcal{T}} p(\tau) = \psi (v_\lambda),
\end{align}
where
\begin{align}
\psi (v) \coloneqq
\begin{cases}
\theta_v (\bm 0) = 1, & v \in \mathcal{L}_\mathrm{p},\\
\max_{\bm z_v \in \{ 0, 1\}^{k_\mathrm{max}}} \Bigl\{ \theta_v(\bm z_v) \\
\qquad \times \prod_{v' \in \mathrm{Ch}_\mathrm{p} (v)} \left( \psi (v') \right)^{z_{vv'}} \Bigr\}, & v \in \mathcal{I}_\mathrm{p}.
\end{cases}
\end{align}
\end{prop}

In addition, we define a flag variable $\bm z^*_v \in \{ 0, 1\}^{k_\mathrm{max}}$ as follows.
\begin{defi}
For any $v \in \mathcal{V}_\mathrm{p}$, we define
\begin{align}
\bm z^*_v = \argmax_{\bm z_v \in \{ 0, 1\}^{k_\mathrm{max}}} \left\{ \theta_v(\bm z_v) \prod_{v' \in \mathrm{Ch}_\mathrm{p} (v)} \left( \psi (v') \right)^{z_{vv'}} \right\}.
\end{align}
\end{defi}

We can calculate $\psi(v)$ and $\bm z^*_v$ simultaneously. Then, the mode of $p(\tau)$ is given by the following proposition.

\begin{prop}
$\mathrm{arg} \max_{\tau \in \mathcal{T}} p(\tau)$ is identified as the tree that satisfies
\begin{align}
v \in \mathcal{V}_\tau \Rightarrow \bm z_v^\tau = \bm z^*_v.
\end{align}
\end{prop}

Then, the following theorem holds.
\begin{theo}\label{mode_alg}
The mode of $p(\tau)$ can be found by backtracking search from $v_\lambda$ after the calculation of $\psi (v)$ and $\bm z_v^*$. It is detailed in Algorithm \ref{calc_mode} in Appendix \ref{pseudocode}.
\end{theo}

\subsection{Expectation}
Let $f: \mathcal{T} \to \mathbb{R}$ denote a real-valued function on $\mathcal{T}$. Here, we discuss sufficient conditions of $f$, under which the following expectation can be calculated with $O \left( 2^{k_\mathrm{max}} k_\mathrm{max}^{d_\mathrm{max}+1} \right)$ cost.
\begin{align}
\mathbb{E} [f(T)] \coloneqq \sum_{\tau \in \mathcal{T}} f(\tau) p(\tau). \label{expectation}
\end{align}
Note that the size of $\mathcal{T}$ is of the order of $\Omega \left( 2^{k_\mathrm{max}^{d_\mathrm{max}-1}} \right)$ in general.

\begin{condi}\label{product}
There exists $g_v: \{ 0, 1\}^{k_\mathrm{max}} \to \mathbb{R}$ for each $v \in \mathcal{V}_\mathrm{p}$ such that
\begin{align}
f(\tau) = \prod_{v \in \mathcal{V}_\tau} g_v(\bm z_v^\tau). \label{product_condition}
\end{align}
\end{condi}

\begin{theo}\label{product-ex}
Under Condition \ref{product}, we define a recursive function $\phi: \mathcal{V}_\mathrm{p} \to \mathbb{R}$ as 
\begin{align}
\phi (v) \coloneqq
\begin{cases}
g_v(\bm 0), & v \in \mathcal{L}_\mathrm{p},\\
\sum_{\bm z_v \in \{ 0, 1\}^{k_\mathrm{max}}} \Bigl\{ \theta_v(\bm z_v) g_v(\bm z_v) \\
\qquad \times \prod_{v' \in \mathrm{Ch}_\mathrm{p} (v)} \left( \phi (v') \right)^{z_{vv'}} \Bigr\}, & v \in \mathcal{I}_\mathrm{p}.
\end{cases}
\end{align}
Then, we can calculate $\mathbb{E} [f(T)]$ as $\mathbb{E} [f(T)] = \phi (v_\lambda)$.
\end{theo}

\begin{exam}
Theorem \ref{node_events} can be regarded examples of Theorem \ref{product-ex}.
\end{exam}

\begin{condi}\label{sum}
There exists $g_v: \{ 0, 1\}^{k_\mathrm{max}} \to \mathbb{R}$ for each $v \in \mathcal{V}_\mathrm{p}$ such that
\begin{align}
f(\tau) = \sum_{v \in \mathcal{V}_\tau} g_v(\bm z_v^\tau). \label{sum_condition}
\end{align}
\end{condi}

\begin{theo}\label{sum-ex}
Under Condition \ref{sum}, we define a recursive function $\xi: \mathcal{V}_\mathrm{p} \to \mathbb{R}$ as 
\begin{align}
\xi (v) \coloneqq
\begin{cases}
g_v(\bm 0), & v \in \mathcal{L}_\mathrm{p},\\
\sum_{\bm z_v \in \{ 0, 1\}^{k_\mathrm{max}}} \Bigl\{ \theta_v (\bm z_v) \Bigl( g_v(\bm z_v)  \\
\qquad \quad + \sum_{v' \in \mathrm{Ch}_\mathrm{p} (v)} z_{vv'} \xi (v') \Bigr) \Bigr\}, & v \in \mathcal{I}_\mathrm{p}.
\end{cases}\label{sum-ex-recursion}
\end{align}
Then, we can calculate $\mathbb{E} [f(T)]$ as $\mathbb{E} [f(T)] = \xi (v_\lambda)$.
\end{theo}

\begin{rema}
Theorem \ref{sum-ex} is useful to calculate the Shannon entropy of $p(\tau)$. It is described in Section \ref{entropy}.
\end{rema}

\subsection{Shannon entropy}\label{entropy}

\begin{coro}\label{theo_entropy}
Substituting $g_v(\bm z) = - \log \theta_v (\bm z)$ into (\ref{sum-ex-recursion}), the Shannon entropy $H[T] \coloneqq - \sum_{\tau \in \mathcal{T}}p(\tau) \log p(\tau)$ can be recursively calculated as follows.
\begin{align}
H[T] = \xi (v_\lambda),
\end{align}
where
\begin{align}
\xi (v) \coloneqq
\begin{cases}
0, & v \in \mathcal{L}_\mathrm{p},\\
\sum_{\bm z_v \in \{ 0, 1\}^{k_\mathrm{max}}} \Bigl\{ \theta_v (\bm z_v) \Bigl( -\log \theta_v(\bm z_v)  \\
\qquad \quad + \sum_{v' \in \mathrm{Ch}_\mathrm{p} (v)} z_{vv'} \xi (v') \Bigr) \Bigr\}, & v \in \mathcal{I}_\mathrm{p}.
\end{cases}
\end{align}
\end{coro}

\begin{rema}
Kullback-Leibler divergence between two tree distributions $p(\tau)$ and $p'(\tau)$ can be calculated in a similar manner to Corollary \ref{theo_entropy}. This fact may be useful for variational Bayesian inference, in which the Kullback-Leibler divergence is minimized. This is a future work.
\end{rema}

\subsection{Conjugate prior of $p(\tau | \bm \theta)$}

Here, we consider that $\theta_v (\bm z) \in [0, 1]$ is also a realization of a random variable. Let $\bm \theta_v$ denote $(\theta_v (\bm z))_{\bm z \in \{ 0, 1\}^{k_\mathrm{max}}}$ and $\bm \theta$ denote $(\bm \theta_v )_{v \in \mathcal{V}_\mathrm{p}}$, and we describe $p(\tau)$ as $p(\tau | \bm \theta)$ to emphasize the dependency of $\bm \theta$ in the following theorem. Then, a conjugate prior for $p(\tau | \bm \theta)$ is as follows.

\begin{theo}\label{conjugate_prior}
The following probability distribution is a conjugate prior for $p(\tau | \bm \theta)$.
\begin{align}
p(\bm \theta) \coloneqq \prod_{v \in \mathcal{V}_\mathrm{p}} \mathrm{Dir} ( \bm \theta_v | \bm \alpha_v),
\end{align}
where $\mathrm{Dir} (\cdot | \bm \alpha_v)$ denotes the probability density function of the Dirichlet distribution whose parameters are $\bm \alpha_v \coloneqq (\alpha_v (\bm z))_{\bm z \in \{ 0, 1\}^{k_\mathrm{max}}} \in \mathbb{R}_{> 0}^{k_\mathrm{max}}$ for each $v \in \mathcal{V}_\mathrm{p}$. More precisely,
\begin{align}
p(\bm \theta | \tau) = \prod_{v \in \mathcal{V}_\mathrm{p}} \mathrm{Dir}(\bm \theta_v | \bm \alpha_{v|\tau}),
\end{align}
where $\bm \alpha_{v|\tau} \coloneqq (\alpha_{v|\tau}(\bm z))_{\bm z \in \{ 0, 1\}^{k_\mathrm{max}}}$ and
\begin{align}
\alpha_{v|\tau} (\bm z) &\coloneqq
\begin{cases}
\alpha_v (\bm z) + 1, & (v \in \mathcal{V}_\tau) \land (\bm z = \bm z_v^\tau), \\
\alpha_v (\bm z), & \mathrm{otherwise}.
\end{cases}
\end{align}
\end{theo}

\subsection{$p(\tau)$ as conjugate prior}
We define another random variable $X$ on a set $\mathcal{X}$ and assume $X$ depends on $T$, i.e., it follows a distribution $p(x | \tau)$. Here, we discuss a sufficient condition of $p(x | \tau)$, under which $p(\tau)$ becomes a conjugate prior for it and we can efficiently calculate the posterior $p(\tau | x)$.

\begin{condi}\label{generative-general}
There exists a function $g_v: \mathcal{X} \times \{ 0, 1\}^{k_\mathrm{max}} \to \mathbb{R}$ for each $v \in \mathcal{V}_\mathrm{p}$, and $p(x | \tau)$ has the following form.
\begin{align}
p(x | \tau) = \prod_{v \in \mathcal{V}_\tau} g_v(x, \bm z_v^\tau). \label{likelihood}
\end{align}
\end{condi}

\begin{theo}\label{posterior_general}
Under Condition \ref{generative-general}, we define $q(x | v)$ and $\theta_v (\bm z_v|x)$ as follows.
\begin{align}
q(x | v) &\coloneqq
\begin{cases}
g_v(x, \bm 0), & v \in \mathcal{L}_\mathrm{p},\\
\sum_{\bm z_v \in \{ 0, 1\}^{k_\mathrm{max}}} \Bigl\{ \theta_v(\bm z_v) g_v(x, \bm z_v) \\
\qquad \times \prod_{v' \in \mathrm{Ch}_\mathrm{p} (v)} \left( \phi (v') \right)^{z_{vv'}} \Bigr\}, & v \in \mathcal{I}_\mathrm{p}.
\end{cases}\label{q_x_v} \\
\theta_v (\bm z_v | x) &\coloneqq
\begin{cases}
\theta_v(\bm z_v), & v \in \mathcal{L}_\mathrm{p}, \\
\frac{\theta_v (\bm z_v) g_v(x, \bm z_v) \prod_{v' \in \mathrm{Ch}_\mathrm{p}(v)} \left( q(x | v') \right)^{z_{v v'}}}{q(x |v)}, & v \in \mathcal{I}_\mathrm{p}.
\end{cases}\label{theta_up_general}
\end{align}
Note that $\theta_v (\bm 0) = 1$ for $v \in \mathcal{L}_\mathrm{p}$ (see Definition \ref{definition_of_distribution}). Then, the posterior $p(\tau | x)$ is represented as follows.
\begin{align}
p(\tau | x) = \prod_{v \in \mathcal{V}_\tau} \theta_v (\bm z_v^\tau | x). \label{posterior_general_eq}
\end{align}
\end{theo}

It should be noted that the calculation of $q(x | v)$ and $\theta_v (\bm z | x)$ requires $O \left( 2^{k_\mathrm{max}} k_\mathrm{max}^{d_\mathrm{max} + 1} \right)$ cost while it requires $\Omega \left( 2^{k_\mathrm{max}^{d_\mathrm{max}-1}} \right)$ cost in general.

Moreover, if we assume the following condition stronger than Condition \ref{generative-general}, we can calculate the posterior $p(\tau | x)$ more efficiently with $O \left( 2^{k_\mathrm{max}} (d_\mathrm{max}+1) \right)$ cost. An example satisfying the following condition will be described in the next section.

\begin{condi}\label{generative-path}
In addition to Condition \ref{generative-general}, we assume that there exists a path from $v_\lambda$ to a leaf node $v_\mathrm{end} \in \mathcal{L}_\mathrm{p}$ and another function $g'_v: \mathcal{X} \times \{ 0, 1\}^{k_\mathrm{max}} \to \mathbb{R}$ for each $v \in \mathcal{V}_\mathrm{p}$, which satisfy
\begin{align}
g_v (x, \bm z) =
\begin{cases}
g'_v (x, \bm z), & (z_{vv_\mathrm{ch}} = 0) \land (v \succeq v_\mathrm{end}), \\
1, & \mathrm{otherwise.}
\end{cases}\label{g_path}
\end{align}
Here, $v_\mathrm{ch}$ is a child node of $v$ on the path from $v_\lambda$ to $v_\mathrm{end}$. In other words, the value of $p(x|\tau)$ is determined by $g_v (x, \bm z)$ for the deepest node on the path from $v_\lambda$ to $v_\mathrm{end}$.
\end{condi}

\begin{coro}\label{posterior-path}
Under Condition \ref{generative-path}, $q(x | v)$ and $\theta_v (\bm z | x)$ are calculated as follows, more efficiently than (\ref{q_x_v}) and (\ref{theta_up_general}).
\begin{align}
&q(x | v) =
\begin{cases}
g_v(x, \bm 0), & v = v_\mathrm{end},\\
\left( \sum_{\bm z_v: z_{vv_\mathrm{ch}} = 0} \theta_v(\bm z_v) g_v(x, \bm z_v) \right) \\
\quad + \left( \sum_{\bm z_v: z_{vv_\mathrm{ch}} = 1} \theta_v(\bm z_v) \right) \phi (v_\mathrm{ch}), & v \succ v_\mathrm{end},\\
1, & \mathrm{otherwise},
\end{cases}\label{q_x_v_path} \\
&\theta_v (\bm z_v | x) =
\begin{cases}
\theta_v(\bm z_v), & v \not\succ v_\mathrm{end}, \\
\frac{\theta_v (\bm z_v) g_v(x, \bm z_v)}{q(x |v)}, & (z_{vv_\mathrm{ch}} = 0) \land (v \succ v_\mathrm{end}),\\
\frac{\theta_v (\bm z_v) q(x | v_\mathrm{ch})}{q(x |v)}, & (z_{vv_\mathrm{ch}} = 1) \land (v \succ v_\mathrm{end}).
\end{cases}\label{theta_up_path}
\end{align}
Note that we need not calculate $q(x | v)$ for $v \not\succeq v_\mathrm{end}$ to update the posterior and it costs only $O \left( 2^{k_\mathrm{max}} (d_\mathrm{max}+1) \right)$.
\end{coro}

\section{Usefulness in statistical decision theory}\label{discussion}

In a similar manner to \cite{full_rooted_tree_arxiv}, our results are useful in model selection and model averaging for the hierarchical model class under the Bayes criterion in statistical decision theory (see, e.g., \cite{Berger}). The proposed probability distribution $p(\tau)$ is a conjugate prior for stochastic models $p(x|\tau)$ satisfying Condition \ref{generative-general} as shown in Theorem \ref{posterior_general}, and the MAP estimate $\mathrm{arg} \max_{\tau \in \mathcal{T}} p(\tau | x)$ can be efficiently calculated by applying Theorem \ref{mode_alg} to the posterior distribution $p(\tau | x)$ obtained by Theorem \ref{posterior_general}. This is the Bayes optimal model selection based on the posterior distribution. That prevents the selection of the deeper tree as mentioned in Section \ref{introduction}.

Furthermore, we can exactly calculate the predictive distribution $\sum_{\tau \in \mathcal{T}} p(x_\mathrm{new} | \tau) p(\tau | x)$ by using Theorem \ref{posterior_general} and Theorem \ref{product-ex} since the stochastic models $p(x|\tau)$ satisfying Condition \ref{generative-general} also satisfy Condition \ref{product}. This is model averaging of all possible trees with Bayes optimal weights. It should be noted that the prior probability of a deep tree, which often corresponds to a complex statistical model, exponentially decays as its depth increases.

Moreover, since the logarithm of a stochastic model $p(x|\tau)$ satisfying Condition \ref{generative-general} satisfies Condition \ref{sum}, we can calculate $\sum_{\tau \in \mathcal{T}} p(\tau | x) \log p(x | \tau)$ by using Theorems \ref{posterior_general} and \ref{sum-ex}. This implies that we can learn hierarchical Bayesian models by variational Bayesian methods (see, e.g., \cite{Bishop}).

\section{Application}\label{application}

\begin{figure}[tbp]
  \centering
  \includegraphics[width=0.45\textwidth]{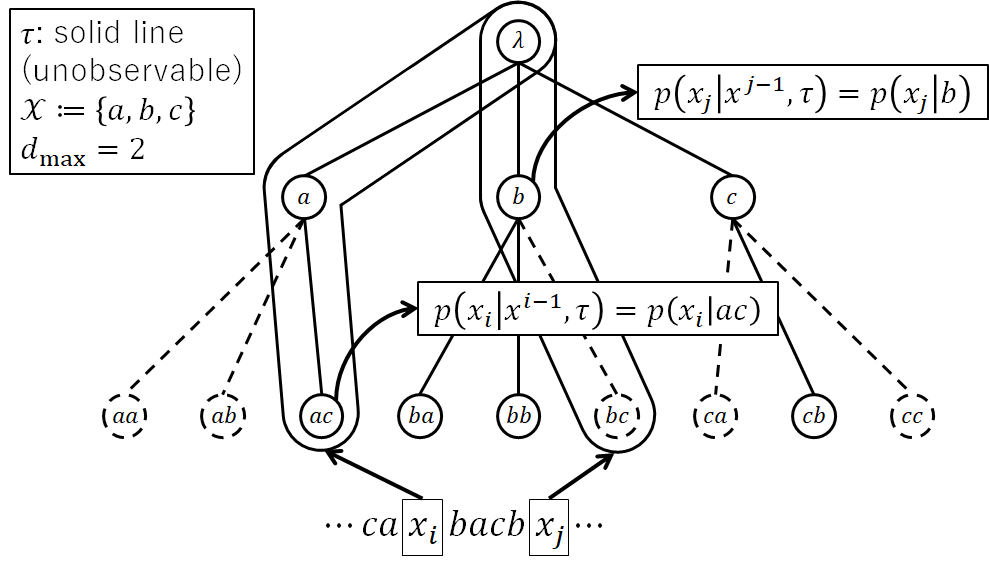}
  \caption{The rooted subtree in text compression (the generalized context tree).}
  \label{CT_model}
\end{figure}

As an example of applications, we generalize the stochastic model assumed in \cite{CT_th, CT_alg, kontoyiannis, Papageorgiou}, which is based on full trees. Let $\mathcal{X}$ denote a source alphabet which consists of $k_\mathrm{max}$ symbols. The node set $\mathcal{V}_\mathrm{p}$ corresponds to the set of contexts, which are strings of symbols shorter than $d_\mathrm{max}$. For example, when $d_\mathrm{max} = 2$, $k_\mathrm{max} = 3$, and $\mathcal{X} = \{ a, b, c \}$, $\mathcal{V}_\mathrm{p} = \{ \lambda, a, b, c, aa, ab, \dots , cc\}$ as shown in Fig.\ \ref{CT_model}. Here, $\lambda$ denotes the empty string. In the previous studies \cite{CT_th, CT_alg, kontoyiannis, Papageorgiou}, the context tree is constructed as a full rooted subtree whose root node is $\lambda$. We generalize it to any rooted subtree $\tau$, which is called generalized context tree herein.

Given past sequence $x^{i-1} \in \mathcal{X}^{i-1}$, a leaf node $v_\mathrm{end}(x^{i-1}) \in \mathcal{L}_\mathrm{p}$ denotes the context $x_{i-1}x_{i-2} \cdots x_{i-d_\mathrm{max}}$ of $x^{i-1}$, whose length is $d_\mathrm{max}$. Then, given generalized context tree $\tau$, next symbol $x_i$ follows the stochastic model\footnote{More precisely, $p(x | v)$ is represented as a compound distribution $\int \mathrm{Cat}(x | \eta_v) \mathrm{Dir}(\eta_v | 1/2, \dots , 1/2) \mathrm{d} \eta_v$, where $\eta_v$ is a parameter of the categorical distribution and follows the Dirichlet prior.} $p(x_i | x^{i-1}, \tau) = p(x_i | v_\tau (x^{i-1}))$ of the longest context $v_\tau (x^{i-1}) \in \mathcal{V}_\tau$ on the path from $\lambda$ to $v_\mathrm{end}(x^{i-1})$. This data generating process is shown in Fig.\ \ref{CT_model}.

Here, if we define $g_v (x_i, \bm z)$ for given $v_\mathrm{end}(x^{i-1})$ as
\begin{align}
g_v (x_i, \bm z) \coloneqq
\begin{cases}
p(x_i | v), & (z_{v v_\mathrm{ch}} = 0) \land (v \succeq v_\mathrm{end} (x^{i-1})), \\
1, & \mathrm{otherwise},
\end{cases}
\end{align}
where $v_\mathrm{ch}$ is a child node of $v$ on the path from $v_\lambda$ to $v_\mathrm{end}(x^{i-1})$, then we have
\begin{align}
p(x_i | x^{i-1}, \tau) &= \prod_{v \in \mathcal{V}_\tau} g_v (x_i, \bm z_v^\tau) \\
&= p(x_i | v_\tau (x^{i-1})).
\end{align}
Therefore, $p(x_i | x^{i-1}, \tau)$ satisfies Condition \ref{generative-path}.

For this model, the optimal coding probability $p^*(x_i | x^{i-1})$ for the arithmetic code can be estimated under the Bayes criterion in statistical decision theory (see, e.g., \cite{Berger}). Such a code is called the Bayes code \cite{Bayes_code}. The coding probability of the Bayes code is given by
\begin{align}
p^*(x_i | x^{i-1}) = \sum_{\tau \in \mathcal{T}} p(\tau | x^{i-1}) p(x_i | x^{i-1}, \tau).
\end{align}
The posterior $p(\tau | x^{i-1})$ can be sequentially calculated by Corollary \ref{posterior-path}, the expectation can be calculated by Theorem \ref{product-ex} or equivalently by (\ref{q_x_v_path}).\footnote{Strictly speaking, $p(x_i | v, x^{i-1}) \coloneqq \int p(x_i | \eta_v, v) p(\eta_v | x^{i-1}, v) \mathrm{d} \eta_v$. Assuming a conjugate prior on $\eta_v$, this can be calculated.}

\begin{rema}\label{small_k_max}
Although the computational cost of our algorithms exponentially increases for $k_\mathrm{max}$, this cost is not so problematic for the case where $k_\mathrm{max}$ is small. For example, we can assume $k_\mathrm{max} = 4$ for genome data since they consists of four alphabets, namely, Adenine, Guanine, Thymine, and Cytosine. In applications other than context tree model, block segmentation of images is represented by quadtrees\cite{nakahara_entropy}, i.e., rooted subtrees with $k_\mathrm{max} = 4$. In the decision tree model\cite{meta-tree} for categorical explanatory variables such as five-level rating, we can assume $k_\mathrm{max} = 5$.
\end{rema}

\begin{figure}[tbp]
  \centering
  \includegraphics[width=0.45\textwidth]{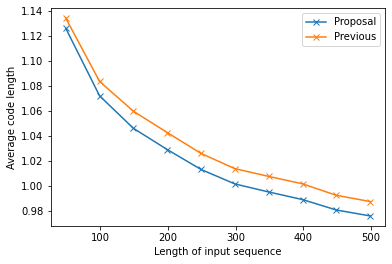}
  \caption{Relation between the average code length and the length of input sequence for our method (blue) and the previous method in \cite{CT_alg} (orange).}
  \label{numerical_result}
\end{figure}

We provide a numerical result of a small experiment for synthetic data in the following. We assumed that $k_\mathrm{max} = 4$, $d_\mathrm{max} = 5$. We generated 100 rooted subtrees according to $p(\tau)$ with $\theta_v (\bm z) \equiv 1/2^{k_\mathrm{max}}$. For each rooted subtree, we generated a sequence $x^n$ according to $\prod_i p(x_i | x^{i-1}, \tau)$.\footnote{Strictly speaking, we generated $(\eta_v)_{v \in \mathcal{V}_\mathrm{p}}$ according to $\prod_{v \in \mathcal{V}_\mathrm{p}} \mathrm{Dir} (\eta_v | 1/2, \dots , 1/2)$ for each sequence, then $x^n$ was generated according to $\prod_i \mathrm{Cat} (x_i | \eta_{v_\tau(x^{i-1})})$.} Then, by using the Bayes code\cite{Bayes_code}, we compressed them by the proposed method and the previous method in \cite{CT_alg},\footnote{We did not compare our method to those in \cite{kontoyiannis, Papageorgiou} since they cannot provide the complete Bayesian inference in contrast to that in \cite{CT_alg}.} which is represented as a specific case of the proposed method in which $\theta_v ((1, 1, \dots , 1)) = 1/2$ and $\theta_v (\bm 0) = 1/2$. Figure \ref{numerical_result} shows the relation between the average code length and the length of input sequence. Our method outperforms the previous method as expected from the Bayes optimality.

\section{Future work}\label{sec-future}
In this study, we did not treat approximative algorithms such as the variational Bayes or Markov chain Monte Carlo method. The proposed algorithms may serve as a subroutine of them. Such algorithms will be useful for  learning hierarchical models that contain the probability distribution on rooted subtrees.

The computational cost of our proposed algorithms exponentially increases for $k_\mathrm{max}$ because they contains the summation with respect to all edge spreading patterns $\bm z_v$. Although there are practical situations where this cost is not so problematic as mentioned in Remark \ref{small_k_max}, it would be serious problem if $k_\mathrm{max}$ is large. To avoid this summation is a meaningful future work.

\section{Conclusion}\label{sec-conclusion}
In this paper, we adopted the theoretical approach to solve the model selection and model averaging for the hierarchical model class represented by rooted trees. On that approach, the rooted trees are regarded as random variables and assumed a parametric prior distribution. In previous studies, such a prior distribution was restricted for full trees, i.e., their inner nodes have the same number of children. In contrast, we proposed a novel prior distribution for general rooted trees. Then, we derived the algorithms that exactly calculate the characteristics of the proposed distribution, namely, the marginal distributions for each node, the mode, the expectations of some class of functions, and the posterior distribution for a class of likelihoods.

\section*{Acknowledgment}

This work was supported by JSPS KAKENHI Grant Numbers JP17K06446, JP19K04914, and JP19K14989.

\bibliographystyle{IEEEtran}
\bibliography{refs_isit}

\appendices

\section{}\label{appendix}

\subsection{Proof of Theorem \ref{node_events}}

Let $I \{ \cdot \}$ denote the indicator function. Then, $\mathrm{Pr} \{ (\bm z_v^T = \bm z) \land (v \in \mathcal{V}_T) \}$ is expressed as
\begin{align}
&\mathrm{Pr} \{ (\bm z_v^T = \bm z) \land (v \in \mathcal{V}_T) \} \nonumber \\
&= \sum_{\tau \in \mathcal{T}} I\{ (\bm z_v^\tau = \bm z) \land (v \in \mathcal{V}_\tau) \} p(\tau).
\end{align}
Here, $v \in \mathcal{V}_\tau$ is equivalent that $z_{v'v''}^\tau = 1$ holds for all the edges $(v', v'')$ composed of the ancestor nodes $v',  v'' \in \mathrm{An}(v) \cup \{ v \}$ of $v$. We define a function $G_{v'; v, \bm z}: \{ 0, 1\}^{k_\mathrm{max}} \to \mathbb{R}$ for each $v' \in \mathcal{V}_\mathrm{p}$, where $v \in \mathcal{V}_\mathrm{p}$ and $\bm z \in \{ 0, 1\}^{k_\mathrm{max}}$ are fixed.
\begin{align}
&G_{v'; v, \bm z} (\bm z_{v'}) \nonumber \\
&\coloneqq
\begin{cases}
1, & v' \not\succeq v, \\
I \{ z_{v' v''} = 1 \text{ for } v'' \succeq v \}, & v' \succ v, \\
I \{ \bm z_{v'} = \bm z \}, & v' = v.
\end{cases}\label{G_v_for_node_events}
\end{align}
Then, since $I \{ (\bm z_v^T = \bm z) \land (v \in \mathcal{V}_T) \} = \prod_{v' \in \mathcal{V}_\tau} G_{v'; v, \bm z} (\bm z_{v'}^\tau)$, the following holds.
\begin{align}
&\mathrm{Pr} \{ (\bm z_v^T = \bm z) \land (v \in \mathcal{V}_T) \} \nonumber \\
&= \sum_{\tau \in \mathcal{T}} \left( \prod_{v' \in \mathcal{V}_\tau} G_{v'; v, \bm z} (\bm z_{v'}^\tau) \right) p(\tau) \\
&= \sum_{\tau \in \mathcal{T}} \prod_{v' \in \mathcal{V}_\tau} \Bigl( G_{v'; v, \bm z} (\bm z_{v'}^\tau) \theta_{v'} (\bm z_{v'}^\tau) \Bigr).
\end{align}

By using Lemma \ref{func_sum}, we have
\begin{align}
\mathrm{Pr} \{ (\bm z_v^T = \bm z) \land (v \in \mathcal{V}_T) \} = \phi_{v, \bm z} (v_\lambda),
\end{align}
where
\begin{align}
\phi_{v, \bm z} (v') \coloneqq
\begin{cases}
G_{v'; v, \bm z}(\bm 0) \theta_{v'} (\bm 0), & v' \in \mathcal{L}_\mathrm{p},\\
\sum_{\bm z_{v'} \in \{ 0, 1\}^{k_\mathrm{max}}} \Bigl\{ \theta_{v'} (\bm z_{v'}) G_{v'; v, \bm z} (\bm z_{v'}) \\
\quad \times \prod_{v'' \in \mathrm{Ch}_\mathrm{p} (v')} \left( \phi_{v, \bm z} (v'') \right)^{z_{v'v''}} \Bigr\}, & v' \in \mathcal{I}_\mathrm{p}.
\end{cases}\label{node_event_recursion}
\end{align}

We further transform this function. We split the remainder into three cases.

\textit{Case 1 ($v' \not\succeq v$):} If $v' \not\succeq v$, then $G_{v'; v, \bm z} (\bm z_{v'}) = 1$ and consequently,
\begin{align}
\phi_{v, \bm z} (v') =
\begin{cases}
\theta_{v'} (\bm 0), & v' \in \mathcal{L}_\mathrm{p}, \\
\sum_{\bm z_{v'} \in \{ 0, 1\}^{k_\mathrm{max}}} \Bigl\{ \theta_{v'} (\bm z_{v'}) \\
\quad \times \prod_{v'' \in \mathrm{Ch}_\mathrm{p} (v')} \left( \phi_{v, \bm z} (v'') \right)^{z_{v'v''}} \Bigr\}, & v' \in \mathcal{I}_\mathrm{p}.
\end{cases}\label{not_on_path}
\end{align}
It has the same form as (\ref{sum_recursive}), and every child node $v'' \in \mathrm{Ch}_\mathrm{p} (v')$ also satisfies $v'' \not\succeq v$. Therefore, $\phi_v (v') = 1$ for $v' \not\succeq v$.

\textit{Case 2 ($v' = v$):} If $v' = v$ and $v \in \mathcal{I}_\mathrm{p}$, then any $v'' \in \mathrm{Ch}_\mathrm{p} (v')$ satisfies $v'' \not\succeq v$. Therefore, we have
\begin{align}
&\phi_{v, \bm z} (v') \nonumber \\
&= \sum_{\bm z_{v'} \in \{ 0, 1\}^{k_\mathrm{max}}} G_{v'; v, \bm z} (\bm z_{v'}) \theta_{v'} (\bm z_{v'}) \prod_{v'' \in \mathrm{Ch}_\mathrm{p} (v')} 1^{z_{v'v''}} \\
&= \theta_{v'} (\bm z),
\end{align}
where the last equation is because of (\ref{G_v_for_node_events}) for $v' = v$. If $v' = v$ and $v \in \mathcal{L}_\mathrm{p}$, then
\begin{align}
\phi_{v, \bm z} (v') &= G_{v'; v, \bm z} (\bm 0) \theta_{v'} (\bm 0) \\
&=
\begin{cases}
1, & \bm z = \bm 0\\
0, & \mathrm{otherwise}
\end{cases}\\
&= \theta_{v'} (\bm z).
\end{align}
Therefore, when $v' = v$, $\phi_{v, \bm z} (v') = \theta_{v'} (\bm z)$ holds for both $v \in \mathcal{I}_\mathrm{p}$ and $v \in \mathcal{L}_\mathrm{p}$.

\textit{Case 3 ($v' \succ v$):} If $v' \succ v$, $v'$ cannot be in $\mathcal{L}_\mathrm{p}$ and has only one child node in $\mathrm{An}(v) \cup \{ v \}$. Let $v'_\mathrm{ch}$ denote it. Then, $\phi_{v, \bm z} (v'') = 1$ for the other child nodes $v'' \in \mathrm{Ch}_\mathrm{p}(v') \setminus \{ v'_\mathrm{ch} \}$. Therefore, (\ref{node_event_recursion}) is represented as follows.
\begin{align}
&\phi_{v, \bm z} (v') \nonumber \\
&= \sum_{\bm z_{v'} \in \{ 0, 1\}^{k_\mathrm{max}}} G_{v'; v, \bm z} (\bm z_{v'}) \theta_{v'} (\bm z_{v'}) \left( \phi_{v, \bm z} (v'_\mathrm{ch}) \right)^{z_{v' v'_\mathrm{ch}}} \\
&= \sum_{\bm z_{v'} \in \{ 0, 1\}^{k_\mathrm{max}}: z_{v'v'_\mathrm{ch}} = 1} \theta_{v'} (\bm z_{v'}) \phi_{v, \bm z} (v'_\mathrm{ch}) \\
&= \phi_{v, \bm z} (v'_\mathrm{ch}) \sum_{\bm z_{v'} \in \{ 0, 1\}^{k_\mathrm{max}}: z_{v'v'_\mathrm{ch}} = 1} \theta_{v'} (\bm z_{v'}).
\end{align}

Therefore, the following holds by expanding $\phi_{v, \bm z} (v_\lambda)$.
\begin{align}
&\mathrm{Pr} \{ (\bm z_v^T = \bm z) \land (v \in \mathcal{V}_T) \} \nonumber \\
&= \theta_v(\bm z) \prod_{(v',v'') \in \mathcal{E}_{v_\lambda \to v}} \sum_{\bm z_{v'}: z_{v'v''} = 1} \theta_{v'} (\bm z_{v'}) .
\end{align}
This completes the proof of Theorem \ref{node_events}. \hfill $\blacksquare$

\subsection{Proof of Theorem \ref{product-ex}}

Substituting (\ref{product_condition}) into (\ref{expectation}), $\mathbb{E}[f(T)]$ can be represented as follows.
\begin{align}
\mathbb{E} [f(T)] = \sum_{\tau \in \mathcal{T}} \prod_{v \in \mathcal{V}_\tau} \theta_v(\bm z_v^\tau) g_v(\bm z_v^\tau).
\end{align}
Then, using Lemma \ref{func_sum}, Theorem \ref{product-ex} straightforwardly follows.

\hfill $\blacksquare$

\subsection{Proof of Theorem \ref{sum-ex}}

First, we switch the order of the summation as follows.
\begin{align}
&\mathbb{E}[f(T)] \nonumber \\
&= \sum_{\tau \in \mathcal{T}} p(\tau) \sum_{v \in \mathcal{V}_\tau} g_v(\bm z_v^\tau) \\
&= \sum_{\tau \in \mathcal{T}} p(\tau) \sum_{v \in \mathcal{V}_\mathrm{p}} \sum_{\bm z_v \in \{ 0, 1\}^{k_\mathrm{max}}} \! I \{ (\bm z_v^\tau = \bm z_v) \! \land \! (v \in \mathcal{V}_\tau) \} g_v(\bm z_v) \\
&= \sum_{v \in \mathcal{V}_\mathrm{p}} \sum_{\bm z_v \in \{ 0, 1\}^{k_\mathrm{max}}} \! g_v(\bm z_v) \sum_{\tau \in \mathcal{T}} p(\tau) I \{ (\bm z_v^\tau = \bm z_v) \! \land \! (v \in \mathcal{V}_\tau) \} \\
&= \sum_{v \in \mathcal{V}_\mathrm{p}} \sum_{\bm z_v \in \{ 0, 1\}^{k_\mathrm{max}}} g_v(\bm z_v) \mathrm{Pr} \{ (\bm z_v^T = \bm z_v) \land (v \in \mathcal{V}_T) \} \\
&= \sum_{v \in \mathcal{V}_\mathrm{p}} \sum_{\bm z_v \in \{ 0, 1\}^{k_\mathrm{max}}} \Biggl\{ g_v(\bm z_v)  \theta_v(\bm z_v) \nonumber \\
& \qquad \qquad \qquad \times \prod_{(v',v'') \in \mathcal{E}_{v_\lambda \to v}} \sum_{\bm z_{v'}: z_{v'v''} = 1} \theta_{v'} (\bm z_{v'}) \Biggr\} \label{using_node_events_theorem} \\
&= \sum_{v \in \mathcal{V}_\mathrm{p}} \Biggl\{ \Biggl( \sum_{\bm z_v \in \{ 0, 1\}^{k_\mathrm{max}}} g_v(\bm z_v)  \theta_v(\bm z_v) \Biggr) \nonumber \\
& \qquad \qquad \qquad \times \prod_{(v',v'') \in \mathcal{E}_{v_\lambda \to v}} \sum_{\bm z_{v'}: z_{v'v''} = 1} \theta_{v'} (\bm z_{v'}) \Biggr\} , \label{sum_ex_before_recursive}
\end{align}
where (\ref{using_node_events_theorem}) is because of Theorem \ref{node_events}.

Next, we decompose the right-hand side of (\ref{sum_ex_before_recursive}) until it has the same form as (\ref{sum-ex-recursion}). 
\begin{align}
(\ref{sum_ex_before_recursive}) &= \sum_{\bm z_{v_\lambda} \in \{ 0, 1\}^{k_\mathrm{max}}} g_{v_\lambda}(\bm z_{v_\lambda})  \theta_{v_\lambda}(\bm z_{v_\lambda}) \nonumber \\
&\qquad + \sum_{v \in \mathrm{De}_\mathrm{p} ( v_\lambda )} \Biggl\{ \Biggl( \sum_{\bm z_v \in \{ 0, 1\}^{k_\mathrm{max}}} g_v(\bm z_v)  \theta_v(\bm z_v) \Biggr) \nonumber \\
& \qquad \qquad \qquad \quad \times \prod_{(v',v'') \in \mathcal{E}_{v_\lambda \to v}} \sum_{\bm z_{v'}: z_{v'v''} = 1} \theta_{v'} (\bm z_{v'}) \Biggr\} \label{decompose1} \\
&= \sum_{\bm z_{v_\lambda} \in \{ 0, 1\}^{k_\mathrm{max}}} g_{v_\lambda}(\bm z_{v_\lambda})  \theta_{v_\lambda}(\bm z_{v_\lambda}) + \sum_{v \in \mathrm{Ch}_\mathrm{p} (v_\lambda)} \Biggl[ \nonumber \\
& \qquad \sum_{v' \in \mathrm{De}_\mathrm{p} (v) \cup \{ v \}} \Biggl\{ \Biggl( \sum_{\bm z_{v'} \in \{ 0, 1\}^{k_\mathrm{max}}} g_{v'}(\bm z_{v'})  \theta_{v'}(\bm z_{v'}) \Biggr) \nonumber \\
& \qquad \qquad \quad \times \prod_{(v'',v''') \in \mathcal{E}_{v_\lambda \to v'}} \sum_{\bm z_{v''}: z_{v''v'''} = 1} \theta_{v''} (\bm z_{v''}) \Biggr\} \Biggr] \label{decompose2} \\
&= \sum_{\bm z_{v_\lambda} \in \{ 0, 1\}^{k_\mathrm{max}}} g_{v_\lambda}(\bm z_{v_\lambda})  \theta_{v_\lambda}(\bm z_{v_\lambda}) \nonumber \\
&\quad + \sum_{v \in \mathrm{Ch}_\mathrm{p} (v_\lambda)} \Biggl( \sum_{\bm z_{v_\lambda}: z_{v_\lambda v} = 1} \theta_{v_\lambda} (\bm z_{v_\lambda}) \Biggr) \nonumber \\
& \quad \times \Biggl[ \sum_{v' \in \mathrm{De}_\mathrm{p} (v) \cup \{ v \}} \Biggl\{ \Biggl( \sum_{\bm z_{v'} \in \{ 0, 1\}^{k_\mathrm{max}}} g_{v'}(\bm z_{v'})  \theta_{v'}(\bm z_{v'}) \Biggr) \nonumber \\
& \qquad \qquad \quad \times \prod_{(v'',v''') \in \mathcal{E}_{v \to v'}} \sum_{\bm z_{v''}: z_{v''v'''} = 1} \theta_{v''} (\bm z_{v''}) \Biggr\} \Biggr], \label{sum_ex_decomposed}
\end{align}
where (\ref{decompose1}) is because $\mathcal{V}_\mathrm{p} = \{ v_\lambda \} \cup \mathrm{De}_\mathrm{p} (v_\lambda)$ and $\mathcal{E}_{v_\lambda \to v_\lambda} = \emptyset$; (\ref{decompose2}) is because $\mathrm{De}_\mathrm{p} (v_\lambda) =  \bigcup_{v \in \mathrm{Ch}_\mathrm{p}(v_\lambda)} ( \mathrm{De}_\mathrm{p} (v) \cup \{ v \} )$; and (\ref{sum_ex_decomposed}) is because $\mathcal{E}_{v_\lambda \to v'}$ contains $(v_\lambda, v)$ and $\mathcal{E}_{v_\lambda \to v'} = \{  (v_\lambda, v) \} \cup \mathcal{E}_{v \to v'}$ for any $v \in \mathrm{Ch}_\mathrm{p} (v_\lambda)$ and $v' \in \mathrm{De}_\mathrm{p}(v) \cup \{ v \}$.

Comparing (\ref{sum_ex_before_recursive}) and (\ref{sum_ex_decomposed}), we have
\begin{align}
&\underbrace{\sum_{v \in \mathcal{V}_\mathrm{p}} \Biggl\{ \Biggl( \sum_{\bm z_v \in \{ 0, 1\}^{k_\mathrm{max}}} g_v(\bm z_v)  \theta_v(\bm z_v) \Biggr) }_{(a)} \nonumber \\
& \qquad \qquad \qquad \underbrace{\times \prod_{(v',v'') \in \mathcal{E}_{v_\lambda \to v}} \sum_{\bm z_{v'}: z_{v'v''} = 1} \theta_{v'} (\bm z_{v'}) \Biggr\} }_{(a)} \\
&= \sum_{\bm z_{v_\lambda} \in \{ 0, 1\}^{k_\mathrm{max}}} g_{v_\lambda}(\bm z_{v_\lambda})  \theta_{v_\lambda}(\bm z_{v_\lambda}) \nonumber \\
&\quad + \sum_{v \in \mathrm{Ch}_\mathrm{p} (v_\lambda)} \Biggl( \sum_{\bm z_{v_\lambda}: z_{v_\lambda v} = 1} \theta_{v_\lambda} (\bm z_{v_\lambda}) \Biggr) \nonumber \\
& \quad \times \underbrace{\Biggl[ \sum_{v' \in \mathrm{De}_\mathrm{p} (v) \cup \{ v \}} \Biggl\{ \Biggl( \sum_{\bm z_{v'} \in \{ 0, 1\}^{k_\mathrm{max}}} g_{v'}(\bm z_{v'})  \theta_{v'}(\bm z_{v'}) \Biggr) }_{(b)} \nonumber \\
& \qquad \qquad \quad \underbrace{\times \prod_{(v'',v''') \in \mathcal{E}_{v \to v'}} \sum_{\bm z_{v''}: z_{v''v'''} = 1} \theta_{v''} (\bm z_{v''}) \Biggr\} \Biggr] }_{(b)} \label{sum_ex_last}
\end{align}
The underbraced parts $(a)$ and $(b)$ have the same structure. Therefore, $(b)$ can be decomposed in a similar manner from (\ref{sum_ex_before_recursive}) to (\ref{sum_ex_decomposed}). We can continue this decomposition to the leaf nodes. 

Finally, we have an alternative definition of $\xi (v): \mathcal{V}_p \to \mathbb{R}$, which is equivalent to (\ref{sum-ex-recursion}).
\begin{align}
\xi (v) \coloneqq & \sum_{v' \in \mathrm{De}_\mathrm{p} (v) \cup \{ v \}} \Biggl\{ \Biggl( \sum_{\bm z_{v'} \in \{ 0, 1\}^{k_\mathrm{max}}} g_{v'}(\bm z_{v'})  \theta_{v'}(\bm z_{v'}) \Biggr) \nonumber \\
& \qquad \qquad \times \prod_{(v'',v''') \in \mathcal{E}_{v \to v'}} \sum_{\bm z_{v''}: z_{v''v'''} = 1} \theta_{v''} (\bm z_{v''}) \Biggr\}. \label{xi_alternative}
\end{align}
The equivalence is confirmed as follows. By substituting (\ref{xi_alternative}) into both sides of (\ref{sum_ex_last}), we have
\begin{align}
&\xi (v_\lambda) \nonumber \\
&= \sum_{\bm z_{v_\lambda} \in \{ 0, 1\}^{k_\mathrm{max}}} g_{v_\lambda}(\bm z_{v_\lambda}) \theta_{v_\lambda}(\bm z_{v_\lambda}) \nonumber \\
&\qquad + \sum_{v \in \mathrm{Ch}_\mathrm{p} (v_\lambda)} \Biggl( \sum_{\bm z_{v_\lambda}: z_{v_\lambda v} = 1} \theta_{v_\lambda} (\bm z_{v_\lambda}) \Biggr) \xi (v) \\
&= \sum_{\bm z_{v_\lambda} \in \{ 0, 1\}^{k_\mathrm{max}}} g_{v_\lambda}(\bm z_{v_\lambda}) \theta_{v_\lambda}(\bm z_{v_\lambda}) \nonumber \\
&\qquad + \sum_{v \in \mathrm{Ch}_\mathrm{p} (v_\lambda)} \Biggl( \sum_{\bm z_{v_\lambda} \in \{ 0, 1 \}^{k_\mathrm{max}}} z_{v_\lambda v} \theta_{v_\lambda} (\bm z_{v_\lambda}) \Biggr) \xi (v) \\
&= \sum_{\bm z_{v_\lambda} \in \{ 0, 1\}^{k_\mathrm{max}}} g_{v_\lambda}(\bm z_{v_\lambda}) \theta_{v_\lambda}(\bm z_{v_\lambda}) \nonumber \\
&\qquad + \sum_{\bm z_{v_\lambda} \in \{ 0, 1 \}^{k_\mathrm{max}}} \sum_{v \in \mathrm{Ch}_\mathrm{p} (v_\lambda)} z_{v_\lambda v} \theta_{v_\lambda} (\bm z_{v_\lambda}) \xi (v) \\
&= \sum_{\bm z_{v_\lambda} \in \{ 0, 1\}^{k_\mathrm{max}}} \theta_{v_\lambda}(\bm z_{v_\lambda}) \Biggl( g_{v_\lambda}(\bm z_{v_\lambda}) + \sum_{v \in \mathrm{Ch}_\mathrm{p} (v_\lambda)} z_{v_\lambda v} \xi (v) \Biggr) .
\end{align}
This completes the proof of Theorem \ref{sum-ex}. \hfill $\blacksquare$

\subsection{Proof of Theorem \ref{conjugate_prior}}

By the Bayes theorem, we have
\begin{align}
p(\bm \theta | \tau) &\propto p(\tau | \bm \theta) p(\bm \theta)\\
&= \prod_{v \in \mathcal{V}_\tau} \theta_v (\bm z_v^\tau) \prod_{v' \in \mathcal{V}_\mathrm{p}} \mathrm{Dir} (\theta_{v'} | \bm \alpha_{v'})\\
&= \prod_{v \in \mathcal{V}_\tau} \theta_v (\bm z_v^\tau) \mathrm{Dir} (\theta_v | \bm \alpha_v) \nonumber \\
&\qquad \times \prod_{v' \in \mathcal{V}_\mathrm{p} \setminus \mathcal{V}_\tau} \mathrm{Dir} (\theta_{v'} | \bm \alpha_{v'}) \\
&\propto \prod_{v \in \mathcal{V}_\mathrm{p}} \mathrm{Dir}(\bm \theta_v | \bm \alpha_{v|\tau}),
\end{align}
where we used the conjugate property between the categorical distribution and the Dirichlet distribution for each term and
\begin{align}
\alpha_{v|\tau} (\bm z) &\coloneqq
\begin{cases}
\alpha_v (\bm z) + 1, & (v \in \mathcal{V}_\tau) \land (\bm z = \bm z_v^\tau), \\
\alpha_v (\bm z), & \mathrm{otherwise}.
\end{cases}
\end{align}
This completes the proof of Theorem \ref{conjugate_prior}.
\hfill $\blacksquare$

\subsection{Proof of Theorem \ref{posterior_general}}

We prove (\ref{posterior_general_eq}) from the right-hand side to the left.
\begin{align}
&\prod_{v \in \mathcal{V}_\tau} \theta_v (\bm z_v^\tau | x) \nonumber \\
&= \prod_{v \in \mathcal{I}_\tau} \theta_v (\bm z_v^\tau | x) \prod_{v' \in \mathcal{L}_\tau \cap \mathcal{L}_\mathrm{p}} \theta_{v'} (\bm z_{v'}^\tau | x) \prod_{v'' \in \mathcal{L}_\tau \cap \mathcal{I}_\mathrm{p}} \theta_{v''} (\bm z_{v''}^\tau | x). \label{three_products}
\end{align}
In the following, we transform each of the above products in order. First, the first product is transformed by substituting (\ref{theta_up_general}) as follows.
\begin{align}
&\prod_{v \in \mathcal{I}_\tau} \theta_v (\bm z_v^\tau | x) \nonumber \\
&= \prod_{v \in \mathcal{I}_\tau} \frac{\theta_v (\bm z_v^\tau) g_v(x, \bm z_v^\tau) \prod_{v' \in \mathrm{Ch}_\mathrm{p}(v)} \left( q(x | v') \right)^{z_{v v'}}}{q(x |v)}. \label{I_p}
\end{align}

Next, the second product is transformed as follows.
\begin{align}
\prod_{v \in \mathcal{L}_\tau \cap \mathcal{L}_\mathrm{p}} \theta_v (\bm z_v^\tau | x) &= \prod_{v \in \mathcal{L}_\tau \cap \mathcal{L}_\mathrm{p}} \theta_v (\bm z_v^\tau) \label{L_p_1}\\
&= \prod_{v \in \mathcal{L}_\tau \cap \mathcal{L}_\mathrm{p}} \frac{\theta_v (\bm z_v^\tau) g_v(x,\bm z_v^\tau)}{q(x | v)}, \label{L_p_2}
\end{align}
where (\ref{L_p_1}) is because of (\ref{theta_up_general}) and (\ref{L_p_2}) is because $q(x | v) = g_v(x,\bm z_v^\tau) = g_v(x,\bm 0)$ for $v \in \mathcal{L}_\mathrm{p}$.

Lastly, the third product is transformed as follows.
\begin{align}
&\prod_{v \in \mathcal{L}_\tau \cap \mathcal{I}_\mathrm{p}} \theta_v (\bm z_v^\tau | x) \nonumber \\
&= \prod_{v \in \mathcal{L}_\tau \cap \mathcal{I}_\mathrm{p}} \frac{\theta_v (\bm z_v^\tau) g_v(x, \bm z_v^\tau) \prod_{v' \in \mathrm{Ch}_\mathrm{p}(v)} \left( q(x | v') \right)^{z_{v v'}^\tau}}{q(x |v)} \label{L_t_1} \\
&= \prod_{v \in \mathcal{L}_\tau \cap \mathcal{I}_\mathrm{p}} \frac{\theta_v (\bm z_v^\tau) g_v(x, \bm z_v^\tau)}{q(x |v)}, \label{L_t_2}
\end{align}
where (\ref{L_t_1}) is because of (\ref{theta_up_general}) and (\ref{L_t_2}) is because $z_{v v'}^\tau = 0$ for $v \in \mathcal{L}_\tau$.

Therefore, we can combine (\ref{I_p}), (\ref{L_p_2}) and (\ref{L_t_2}). Then,
\begin{align}
(\ref{three_products}) &= \prod_{v \in \mathcal{I}_\tau} \frac{\theta_v (\bm z_v^\tau) g_v(x, \bm z_v^\tau) \prod_{v' \in \mathrm{Ch}_\mathrm{p}(v)} \left( q(x | v') \right)^{z_{v v'}}}{q(x |v)} \nonumber \\
&\qquad \times \prod_{v' \in \mathcal{L}_\tau} \frac{\theta_v (\bm z_v^\tau) g_v(x, \bm z_v^\tau)}{q(x |v)}. \label{telescope}
\end{align}
Here, (\ref{telescope}) is a telescoping product, i.e., $q(x|v)$ appears at once in each of the denominator and the numerator. Therefore, we can cancel them except for $q(x | v_\lambda)$. Then,
\begin{align}
(\ref{telescope}) &= \frac{1}{q(x | v_\lambda)} \prod_{v \in \mathcal{V}_\tau} \theta_v (\bm z_v^\tau) g_v(x, \bm z_v^\tau) \\
&= \frac{1}{q(x | v_\lambda)} \prod_{v \in \mathcal{V}_\tau} \theta_v (\bm z_v^\tau) \prod_{v \in \mathcal{V}_\tau} g_v(x, \bm z_v^\tau) \\
&= \frac{p(\tau)p(x | \tau)}{q(x | v_\lambda)}, \label{bayes_theorem_form}
\end{align}
where we used (\ref{likelihood}) and Definition \ref{definition_of_distribution}.

In addition, because of Theorem \ref{product-ex},
\begin{align}
q(x | v_\lambda) = \mathbb{E} [p(x | T)] = \sum_{\tau \in \mathcal{T}} p(x | \tau) p(\tau) = p(x). \label{marginal}
\end{align}
Therefore, 
\begin{align}
(\ref{bayes_theorem_form}) = \frac{p(x|\tau)p(\tau)}{p(x)} = p(\tau|x).
\end{align}
Then, Theorem \ref{posterior_general} holds. \hfill $\blacksquare$

\subsection{Proof of Corollary \ref{posterior-path}}

We will prove only $q(x|v) = 1$ for $v \not\succeq v_\mathrm{end}$. Then, (\ref{q_x_v_path}) and (\ref{theta_up_path}) are straightforwardly derived by substituting it with (\ref{g_path}) into (\ref{q_x_v}) and (\ref{theta_up_general}).

For $v \not\succeq v_\mathrm{end}$, substituting (\ref{g_path}) into (\ref{q_x_v}),
\begin{align}
q(x | v) =
\begin{cases}
1, & v \in \mathcal{L}_\mathrm{p},\\
\sum_{\bm z_v \in \{ 0, 1\}^{k_\mathrm{max}}} \Bigl\{ \theta_v(\bm z_v) \\
\qquad \times \prod_{v' \in \mathrm{Ch}_\mathrm{p} (v)} \left( \phi (v') \right)^{z_{vv'}} \Bigr\}, & v \in \mathcal{I}_\mathrm{p}.
\end{cases}
\end{align}
Since this has the same form as (\ref{not_on_path}), $q(x|v) = 1$ for $v \not\succeq v_\mathrm{end}$ is derived in a similar manner. \hfill $\blacksquare$

\newpage
\section{Pseudocode to calculate mode of $p(\tau)$}\label{pseudocode}
\begin{figure}[h]
\begin{algorithm}[H]
\caption{Calculation of mode of $p(\tau)$}
\label{calc_mode}
\begin{algorithmic}[1]
\Require $\{ \theta_v (\bm z) \}_{(v, \bm z) \in \mathcal{V}_\mathrm{p} \times \{ 0, 1\}^{k_\mathrm{max}}}$
\Ensure $\tau^* = \mathrm{arg} \max_\tau p(\tau)$
\Function {flag\_calculation}{$v$} \Comment {Subroutine}
  \If {$v \in \mathcal{L}_\mathrm{p}$}
    \State $\bm z^*_v \leftarrow \bm 0$
    \State \Return 1
  \ElsIf {$v \in \mathcal{I}_\mathrm{p}$}
    \ForAll {$v' \in \mathrm{Ch}_\mathrm{p}(v)$}
      \State $\mathtt{tmp}(v') \leftarrow \Call {flag\_calculation}{v'}$
    \EndFor
    \State $\bm z^*_v \leftarrow \mathrm{arg}\max_{\bm z_v} \Bigl\{ \theta_v(\bm z_v)$ \\
    \hfill $\times \prod_{v' \in \mathrm{Ch}_\mathrm{p} (v)} \left( \mathtt{tmp}(v') \right)^{z_{vv'}} \Bigr\} \quad$
    \State \Return $\theta_v(\bm z^*_v) \prod_{v' \in \mathrm{Ch}_\mathrm{p} (v)} \left( \mathtt{tmp}(v') \right)^{z^*_{vv'}}$
  \EndIf
\EndFunction
\State
\Function {backtracking}{$v, \mathcal{V}, \mathcal{E}$} \Comment {Subroutine}
  \ForAll {$v' \in \mathrm{Ch}_\mathrm{p}(v)$}
    \If {$z^*_{vv'} = 1$}
      \State $\mathcal{V} \leftarrow \mathcal{V} \cup \{ v' \}$
      \State $\mathcal{E} \leftarrow \mathcal{E} \cup (v, v')$
      \State \Call {backtracking}{$v', \mathcal{V}, \mathcal{E}$}
    \EndIf
  \EndFor
  \State \Return
\EndFunction
\State
\Procedure {}{} \Comment {The main procedure}
\State \Call {flag\_calculation}{$v_\lambda$}
\State $\mathcal{V} \leftarrow \emptyset$
\State $\mathcal{E} \leftarrow \emptyset$
\State \Call {backtracking}{$v_\lambda, \mathcal{V}, \mathcal{E}$}
\State \Return $\tau^* = (\mathcal{V}, \mathcal{E})$
\EndProcedure
\end{algorithmic}
\end{algorithm}
\end{figure}

\end{document}